\documentclass[acmtog]{acmart}

\usepackage{url}
\usepackage{multirow}
\usepackage{listings}
\usepackage[ruled]{algorithm2e}
\usepackage{makecell}
\usepackage{pifont}
\usepackage{color}
\usepackage{hyperref}

\AtBeginDocument{%
  \providecommand\BibTeX{{%
    \normalfont B\kern-0.5em{\scshape i\kern-0.25em b}\kern-0.8em\TeX}}}

\setcopyright{acmcopyright}
\copyrightyear{2021}
\acmYear{2021}
\acmDOI{10.1145/3474088}

\acmJournal{TOG}
\acmVolume{37}
\acmNumber{4}
\acmArticle{15}
\acmMonth{6}

\acmSubmissionID{TOG-20-0085}

\begin{document}

\title{Reliable Image Dehazing by NeRF}

\author{Zheyan Jin}
\orcid{0000-0001-8466-7520}
\email{11930051@zju.edu.cn}

\affiliation{%
  \institution{Zhejiang University}
  \department{State Key Laboratory of Modern Optical Instrumentation}
  \city{Hangzhou}
  \state{Zhejiang}
  \country{China}
}

\author{Huajun Feng}
\email{fenghj@zju.edu.cn}
\affiliation{%
  \institution{Zhejiang University}
  \department{State Key Laboratory of Modern Optical Instrumentation}
  \city{Hangzhou}
  \state{Zhejiang}
  \country{China}
}

\author{Shiyi Chen}
\email{chensq@zju.edu.cn}
\affiliation{%
  \institution{Zhejiang University}
  \department{State Key Laboratory of Modern Optical Instrumentation}
  \city{Hangzhou}
  \state{Zhejiang}
  \country{China}
}

\author{Zhihai Xu}
\email{xuzhh@zju.edu.cn}
\affiliation{%
  \institution{Zhejiang University}
  \department{State Key Laboratory of Modern Optical Instrumentation}
  \city{Hangzhou}
  \state{Zhejiang}
  \country{China}
}

\author{Qi Li}
\email{liqi@zju.edu.cn}
\affiliation{%
  \institution{Zhejiang University}
  \department{State Key Laboratory of Modern Optical Instrumentation}
  \city{Hangzhou}
  \state{Zhejiang}
  \country{China}
}

\author{Yueting Chen}
\orcid{0000-0002-2759-9784}
\email{chenyt@zju.edu.cn}
\affiliation{%
  \institution{Zhejiang University}
  \department{State Key Laboratory of Modern Optical Instrumentation}
  \city{Hangzhou}
  \state{Zhejiang}
  \country{China}
}

\authorsaddresses{%
Authors’ addresses: Z. jin, S. chen, H. Feng, Z. Xu, Q. Li, and Y. Chen, Zhejiang University; emails: 11930051@zju.edu.cn, chenshiqi@zju.edu.cn, fenghj@zju.edu.cn, xuzh@zju.edu.cn, liqi@zju.edu.cn, chenyt@zju.edu.cn}
\renewcommand{\shortauthors}{Chen, et al.}
\begin{abstract}
We present an image dehazing algorithm with high quality, wide application, and no data training or prior needed. We analyze the defects of the original dehazing model, and propose a new and reliable dehazing reconstruction and dehazing model based on the combination of optical scattering model and computer graphics lighting rendering model. Based on the new haze model and the images obtained by the cameras, we can reconstruct the three-dimensional space, accurately calculate the objects and haze in the space, and use the transparency relationship of haze to perform accurate haze removal. To obtain a 3D simulation dataset we used the Unreal 5 computer graphics rendering engine. In order to obtain real shot data in different scenes, we used fog generators, array cameras, mobile phones, underwater cameras and drones to obtain haze data. We use formula derivation, simulation data set and real shot data set result experimental results to prove the feasibility of the new method. Compared with various other methods, we are far ahead in terms of calculation indicators (4 dB higher quality average scene), color remains more natural, and the algorithm is more robust in different scenarios and best in the subjective perception. 

\end{abstract}



\maketitle

\section{Introduction}

Image dehazing is used to deal with the negative effects of haze scattering on imaging. Without a suitable and good image dehazing algorithm, tasks such as image signal processing and image recognition will fail. Image dehazing has been developing rapidly for more than a decade, but there are still several problems.
\begin{figure}[h]
  \centering
  \includegraphics[width=\linewidth]{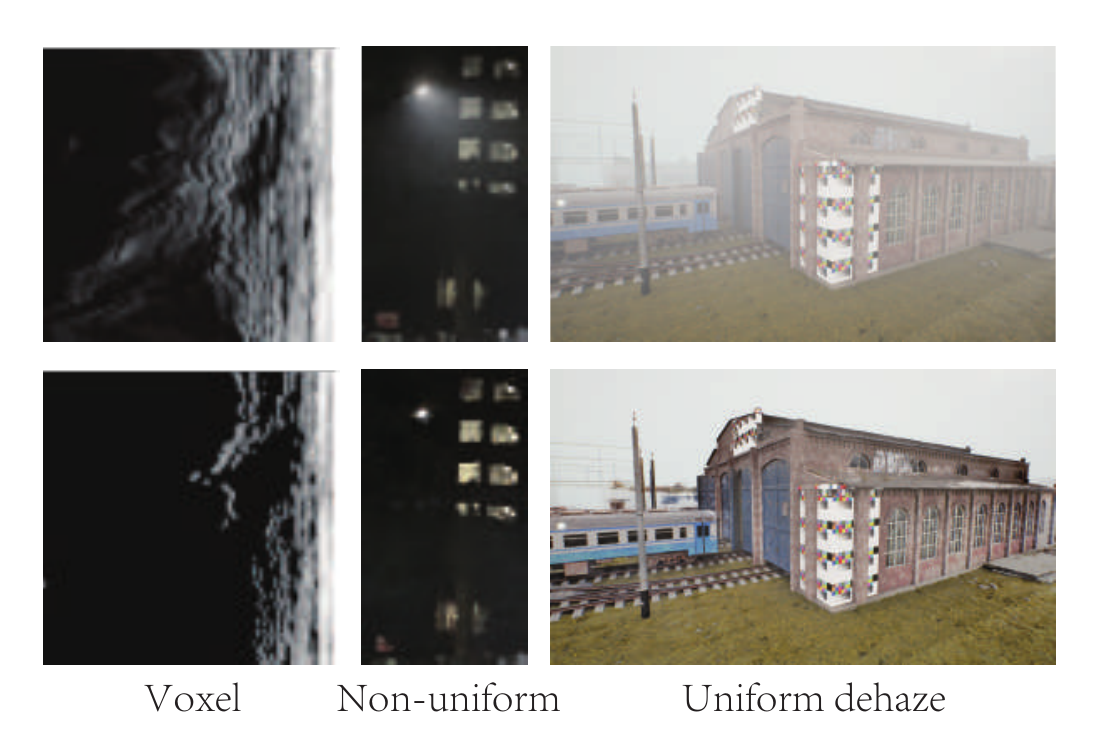}
  \caption{ \textbf{The relationship between voxels and haze.} We create semi-transparent voxels in space from the image's haze. Remove haze by removing voxels. The formula derivation, simulation experiments, and real-shot experiments are used to demonstrate the method's effectiveness and dependability. Because of this, we can reliably and accurately remove haze in a variety of situations without the need for extensive data training or complicated prior knowledge.}
  \Description{A woman and a girl in white dresses sit in an open car.}
  \label{fig_p0}
\end{figure}
%

Pool stability of the image hazing algorithm. The existing algorithms are very good at fitting image data pairs, but it is difficult to make suitable processing for non-uniform haze scenes that have never occurred in dataset. And the haze concentration in many scenes will change, dehazing algorithm is needed to ensure that haze-free scenes can also be kept intact after dehazing.

Poor generalization of image defogging algorithms. Existing dehazing scenes are so much, such as common ground dehazing, remote sensing dehazing, night dehazing, underwater dehazing, sand and dust removal, rain and snow removal. Different scenes also have multiple haze coexist and subdivision changes, so it is time-consuming and ineffective to study a priori information for each type to method.

Image dehazing have little data and are expensive to produce. Many image problems such as de-mosaicing, de-blurring, de-purpling, de-jittering, de-noising, \textit{etc.} due to the components that cause the problems are inside the camera. It is relatively easy to perform internal camera calibration and dataset acquisition. Haze data is often outdoors and the lighting and haze conditions are difficult to keep same.

So we want to solve the above problems once by our new method. In addition to its excellent dehazing results, our method can also bring new idea to the image dehazing. Improve the performance and data quality of various types of image dehazing algorithms. Lays the groundwork for building 3D scenes in a variety of harsh climates.

We propose a lightting scattering model formula that all light scattering conforms to, combining it with the rendering formula. The equivalence of the two is proved by the derivation and experiment. The steady state of the haze scene can be changed into the rendering formula and reconstructed according to the rendering formula. 
After that we create a 3D spatial radiation field to calculate the haze color and intensity in the steady state of haze scattering for each region in space. Translucent voxels in space are created from the haze of the image. The haze is removed by removing the voxels. The dehazing image is obtained by re-rendering the hazing voxels-free space.
The validity and reliability of our method is demonstrated by equation derivation, simulation experiments and real-world photography experiments. Because of this, we can remove haze reliably and accurately in various situations without extensive data training or complex prior knowledge.

Our main technical contributions are as follows:
\begin{itemize}
 
\item  Abandon simple dehazing formulas and use more detailed lightting scattering formulas. Combine the scattering formula with the 3D reconstruction through the computer graphics rendering formula.The calculation principle is more in line with the real world, so dehazing result is more reliable and better.

\item The algorithm can be used in many fields such as ordinary dehazing, nighttime fog dehazing, remote sensing dehazing and underwater dehazing, and does not require the original prior knowledge of each field.

\item The real-shot dehazing datasets are difficult to collect in previous. Since our method does not require a prior knowledge and corresponding image pairs to train, use our method, a large number of multi-scene real-shot datasets can be produced. 

\end{itemize}

\section{related work}
 
\subsection{Image Dehazing}
Image dehazing is a type of low-level computing vision image restoration. Tang \cite{2} et al. used random forest regressor to estimate haze, randomly sampled from multiple clean images extracted various multi-scale features related to fog and then synthesized fog maps. The experimental results once again proved the importance of the dark channel feature DCP \cite{1}, and prove that the integration of various features can more accurately estimate the degree of cloud and fog coverage.
The method of dehazing based on deep learning can be divided into two stages: the initial network training to get intermediate parameters and then substitute the atmospheric degradation model to calculate the final fog-free image. Later models tend to directly learn from foggy images to fog free. The mapping of the fog image (called end-to-end) omits the solving of intermediate parameters, thereby reducing the generation of errors. In 2016, Cai \cite{3} et al introduced an end-to-end CNN network called DehazeNet. The input of the model is a contaminated foggy image, and the output is the transmittance map $t(x)$ of the entire image, and then $t(x)$ and estimated the global atmospheric light is substituted into the degradation model to calculate a clean defogging image. Ren \cite{4} et al proposed a multi-scale deep neural network to estimate the transmittance. The limitation of these methods is that only the transmittance is estimated separately through the CNN framework, so that the errors are amplified by each other. Chen \cite{5} et al proposed a threshold fusion sub-network, which uses GAN to achieve image defogging, and solves the common problem of unreal ghosting. With the recent development of transformer network architecture, DehazeFormer \cite{dehazeformer} be the SOTA of publicly defogged data SOTS \cite{2018RESIDE}.

Compared with ordinary image dehazing, the scene conditions for image dehazing at night are more complicated, and the reasearch started relatively late. Jing, Z et al proposed the NDIM \cite{NDIM} algorithm, which process a color correction step after estimating the color characteristics of the incident light. Li \cite{NHRG} et al. distinguished atmospheric light, haze light, glow, and light sources of different colors, and proposed an NHRG algorithm based on special processing of glow and recognition of different light sources at night. Ancuti \cite{nighttime2016} et al proposed a multi-scale patched pyramid network for artificial light sources to fit the night haze environment. It is believed that the local maximum intensity of each color channel of the night image is mainly contributed by the ambient lighting, and the priori of the maximum reflectance is proposed, and the MRP \cite{MRP} algorithm is designed.Later, the team also proposed a new method of constructing foggy data at night called OSFD \cite{OSFD}, based on scene geometry, and then two-dimensional simulation of light and object reflectivity. They use the new haze rendered image proposed new algorithm and a benchmark test method.

Image defogging algorithms have also evolved into the field of video , which often use multiple frames of information between the temporal variation of the object and haze \cite{EVDnet}.

\subsection{View Render}

Neural Radiance Fields (NeRF) \cite{NeRF} is an implicit MLP-based model that maps 5D vectors—3D coordinates plus 2D viewing directions—to opacity and color values, computed by fitting the model to a set of training views.
NeRF++ \cite{2020NeRF}tries to solve the ambiguity problem of image reconstruction in NeRF, and present a novel spatial parameterization scheme. PixelNeRF \cite{pixelNeRF} can be achieved with fewer images. NGP \cite{NGP} use of hash coding and other acceleration methods greatly speeds up the computation of neural radiation fields. 

In addition to the reduction in the number of images required and calculation overhead, there are variations of NeRF type algorithms for each area. BlockNeRF \cite{BlockNeRF} which focuses on street view generation, megaNeRF \cite{MegaNeRF} which focuses on large scale images, wildNeRF \cite{wildNeRF} which focuses on image fusion in different exposure ranges, and darkNeRF \cite{darkNeRF} which focuses on low illumination at night.

\subsection{Haze Image Dataset and Rending}
The most famous image dehazing data set is RESIDE \cite{2018RESIDE}. It is divided into ITS indoor data set, OTS outdoor data set, HSTS mixed subjective test set, and SOTS comprehensive subjective test set. There is also a dehazing dataset synthesized by NYU2Depth \cite{NYU}. Both of these data sets are simulated. The CVPR NTIRE workshop from 2018 to 2021  released a data set for competition every year, such as O-HAZE \cite{2018O} and I-HAZE \cite{2018I} data sets, which are outdoor and indoor real shot data sets. Later, the outdoor real-shot non-uniform defogging image pair datasets DenseHaze \cite{2019Dense} and NH-HAZE \cite{2020NH-HAZE} were released. But the amount of these data is relatively small, the scene is relatively single, there is still a gap between the number of training needs.

The defogging data generation is divided into two types: mask construction and physical prior rendering. The mask construction is generally based on the haze atmospheric transmission model, which overlays the haze on the fog-free image, such as the RICE data set \cite{2021RICE}. The other such as OSFD \cite{OSFD} which divides the scene in the middle of the image semantically and then performs lighting and texture re-rendering based on the physical model.

\section{METHODOLOGY}

In this work, we build a three-dimensional space to address the issue of image dehazing. A three-dimensional volume in reality corresponds to each neural radiation field voxel. We can obtain a haze-free image by removing the voxels that represent the haze from the generated neural radiation field and then re-rendering.

This approach can eliminate translucent voxels in space. It can be used for image dehazing, specular de-reflection, night dehazing, remote sensing dehazing, underwater dehazing and other applications.

In the subsections that follow, we first give a description of the new optical haze scattering model and contrast it with the optical dehazing and computational graphics fog rendering models that are already in use. Second, we describe the connection between the 3D neural radiance field and the haze space construction. It is explained how non-transparent voxels and haze particles are related. Then, to address voxel thresholding in neural radiation fields, we introduce an image constraint. We discuss how variations in the application of various haze types because of lighting factors next. The primary process of the entire algorithm introduced at last.

\subsection{More Reliable Haze Model}
Image degradation caused by haze is mainly due to the scattering of light by particles. However, the lighting of the actual scene is not always even and consistent. Haze appear differently in nighttime images for various reasons. There is a uniform illumination of atmospheric air light globally. Direct light is fog that is directly illuminated, while glow is the halo haze near the light. These three types of light with different colors and intensities travel to the CMOS through scattering finally. 
Therefore, we must review and enhance the original dehazing model.

The traditional defogging formula is as Eq~\ref{eq1}:

\begin{equation}
I(x) = R(x)*t(x)+L(x)*(1-t(x)) 
\label{eq1}
\end{equation}
where $x$ is the position of the pixel, $I(x)$ is the signal received by the camera pixel, $R(x)$ is the signal emitted by the object itself, $L(x)$ is the atmospheric global illumination, and $t(x)$ is a transmission rate. The transmission rate formula as Eq~\ref{eq2}:
\begin{equation}
t(x)=e^{-\beta \cdot d(x)}
\label{eq2}
\end{equation}
where $d(x)$ is the distance from the object to the camera,    $\beta$ is the attenuation coefficient, and $e$ shows that the attenuation is exponentially linear.

The aforementioned formula introduces depth information, but it is wrong that haze-induced light scattering does simply in the depth direction. This model's dehazing frequently departs greatly from reality. Due to its complexity, light scattering should be thinken into three dimension space.

Back to the original principle of haze formation. The formation of haze is mainly divided into the absorption, out-scattering, emission, in-scattering.

Absorption refers to the light intensity absorbed by the haze particles as Eq~\ref{eqabsorption}. $x$ represents haze particles and $\omega$ represents the angle which light emerges from the haze particles. $\sigma_{a}(x)$ represents absorption coefficient.

\begin{equation}
d L(x, \omega) / d x=-\sigma_{a} L(x, \omega)
\label{eqabsorption}
\end{equation}

Radiative transfer equation (RTE) as Eq ~\ref{eqRTE1} and as Eq ~\ref{eqRTE2}. $-\sigma_{s} L(x, \omega)$ represents out-scattering. $\sigma_{s}(x)$ represents out-scattering coefficient. $\sigma_{a} L_{e}(x, \omega)$ represents emission and  $ f_{p}\left(x, \omega, \omega^{\prime}\right)$ is a phase function. $\int_{s^{2}} f_{p}\left(x, \omega, \omega^{\prime}\right) L\left(x, \omega^{\prime}\right) d \omega^{\prime}$ represents in-scattering.

\begin{equation}
\begin{split}
d L(x, \omega) / d x=-\sigma_{t} L(x, \omega)+\sigma_{a} L_{e}(x, \omega)\\
+\sigma_{s} \int_{s^{2}} f_{p}\left(x, \omega, \omega^{\prime}\right) L\left(x, \omega^{\prime}\right) d \omega^{\prime}
\end{split}
\label{eqRTE1}
\end{equation}

\begin{equation}
 \sigma_{t}(x)=\sigma_{a}(x)+\sigma_{s}(x)
\label{eqRTE2}
\end{equation}

Approximate the derivation of the above formula to the volume can be expressed as ~\ref{eqabsorption}, we call it volume rendering equation (VRE).  $M$ is opaque surface. $L_{i}(x, \omega)$ is in scattering in \ref{eqRTE2}. Transmittance is the net reduction factor from absorption and out-scattering as Eq.\ref{eqT}. 


\begin{equation}
\begin{split}
L(P, \omega)=\int_{x=0}^{d} T(x)\left[\sigma_{a} \cdot L_{e}(x, \omega)+\sigma_{s} \cdot L_{i}(x, \omega)\right] dx \\ +T(M) L(M, \omega)
\end{split}
\label{eqVRE}
\end{equation}   

\begin{equation}
 T(x)=e^{-\int_{x}^{p} \sigma_{t}(s) d s}
\label{eqT}
\end{equation}

Therefore, the above RTE and VRE formulas are the real formulas for light scattering in the real world and computer graphics (CG) rendering. If you want to pursue accurate image dehazing, you must abandon the simple dehazing formula as shown in Eq.~\ref{eq1} and Eq.~\ref{eq2}, and use RTE or VRE instead.
Unfortunately, the existing image processing technology is difficult to deal with this three-dimensional lighting problem. But computer graphics rendering has a very in-depth study on the fog effect.

\begin{figure*}
	\centering
	
    \centerline{\includegraphics[width=\textwidth]{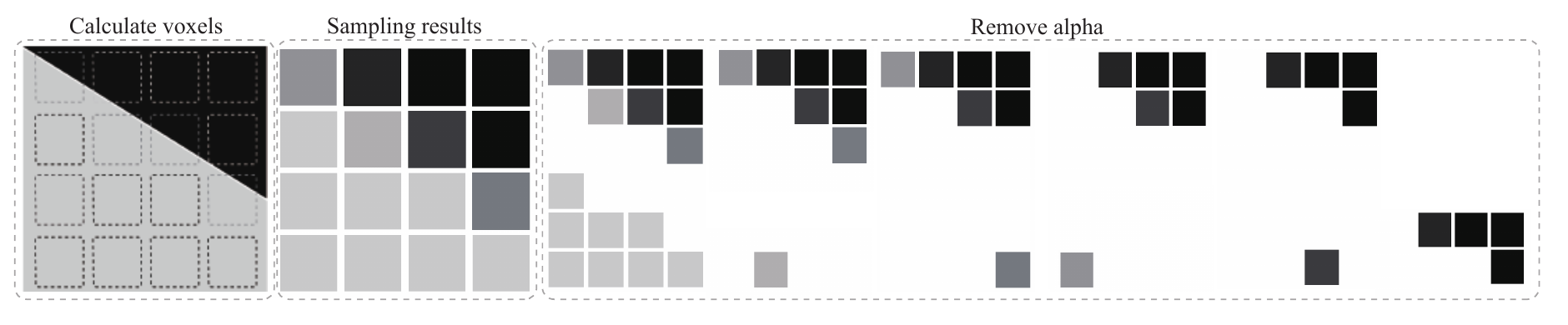}}
    \caption{\textbf{The voxel threshold removal's regularity}There are specific guidelines for voxel removal that are based on the haze and object distribution found in real life.
    Not all voxels are objects or haze due to the accuracy of sampling,  some voxels are frequently sampled to the boundaries of haze and objects.  As seen, there are some voxels of intermediate grey voxels among the mostly light grey and black voxels. However, for the entire voxel space, this portion of the boundary is quite small.There will be three phases when the threshold starts to change: the removal of a large amount of haze, the removal of a small amount of haze and a small amount of object details, and the removal of a large amount of object details. So long as we capture the threshold in the middle phases, we can successfully remove the haze and leaves a significant amount of object details. }
    \label{fig_p1bestssim}
\end{figure*}

\subsection{Rebuild Haze Volume}
The above formulas for RTE and VRE are very complex, and require not only a large amount of computational overhead, but also a large number of parameter acquisition processes. It is difficult to do with image information alone.
Production quick sky and atmosphere technology can reduce the pressure of haze reconstruction.
All points within the neighborhood of the position we currently shade receive the same amount of second order scattered light. Scattered light is $G_{n}$ and currently shade receive is $ {f}_{m s}$ as in Eq.~\ref{eqGn+1}:

\begin{equation}
 G_{n+1}=G_{n} * \boldsymbol{f}_{m s}
\label{eqGn+1}
\end{equation}

Different order scattered light can be approximated as a linear relationship as in Eq ~\ref{eqF}:

\begin{equation}
\mathbf{F}_{\mathrm{ms}}=1+\mathbf{f}_{\mathrm{ms}}+\mathbf{f}_{\mathrm{ms}}^{2}+\mathbf{f}_{\mathrm{ms}}^{3}+\ldots=\frac{1}{1-\mathbf{f}_{\mathrm{ms}}}
\label{eqF}
\end{equation}

It is proved that the final result of light scattering must enter a convergent steady state. That is to say, when we start to calculate the haze in the space, we do not need to calculate the complex VRE process of back and forth scattering of light, but only need to calculate the steady state.

\subsection{Voxel 3D Haze Reconstruction}
After analyzing the real haze scattering model, rendering the model and the scattering steady state, let's take a look at the 3D reconstruction model we used to fit this haze scattering steady state. $\boldsymbol{o}$ is the 3D position of the camera pixel. $\boldsymbol{d}$ is the three-dimensional vector of the direction of the light emitted by the camera.

\begin{equation}
\boldsymbol{r}(t)=\boldsymbol{o}+t \boldsymbol{d}
\label{eqot}
\end{equation}

\begin{equation}
\boldsymbol{C}=(r, g, b)=\int_{t_{\eta}}^{t_{f}} T(t) \sigma(\boldsymbol{r}(t)) \boldsymbol{c}(\boldsymbol{r}(t), \boldsymbol{d}) d t
\label{eqC}
\end{equation}

\begin{equation}
T(t)=\exp \left(-\int_{t_{n}}^{t} \sigma(\boldsymbol{r}(s)) d s\right)
\label{eqtx}
\end{equation}
$T(t)$ is the cumulative transparency of rays. $\boldsymbol{C}$ is the color that goes into the camera at the end.
Then the calculation formula of the whole 3D space can be expressed as Eq.\ref{eqnerf}. Completed the conversion of image to three-dimensional space.
\begin{equation}
F_{MLP}:[(x, y, z),(\theta, \psi) \rightarrow[(r, g, b), \alpha]
\label{eqnerf}
\end{equation}

\begin{figure*}
	\centering
    \centerline{\includegraphics[width=1\textwidth]{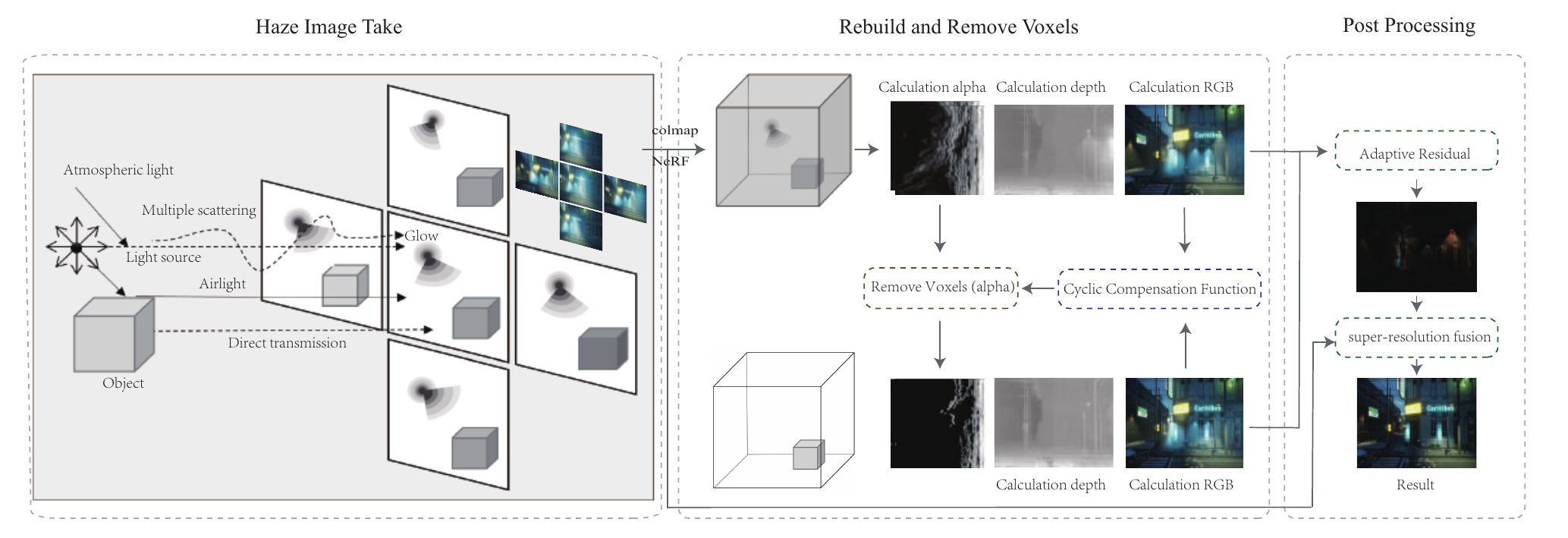}}
    \caption{\textbf{System overview.} We developed a hardware and software-integrated haze removal system. Haze Image Take, Rebuild and Remove Voxels, and Post Processing are the three modules that make up the system.
    Either our camera array or the movement of  a single camera to capture images of the haze. Then, using a number of images, the hazy 3D voxel space is recreated. We also choose an appropriate threshold to eliminate the voxels that are thought to be hazy in this area. By using the image's change information, we limit the threshold. To obtain the required haze-free image, we re-render the 3D space with the haze voxels removed. In order to enhance the quality of rendered haze-free images due to various haze causes and information loss during rendering, we select specific image post-processing techniques.
    Image post-processing including image super-resolution and dehazing optimization for uniform and non-uniform haze scenes.
    Because our image dahazing processing is more accurate and closely resembles the real haze scene, it produces processing results that are more precise, the algorithm is more reliable, and it does not need prior knowledge or data-driven training.}
    \label{fig_p3pipe}
\end{figure*}

\subsection{Voxels and Volumetric fog}

At the same time, due to the existence of  Equation~\ref{eqF}, the multi-level scattering can be approximated as one time, Equation \ref{eqVRE} and Equation \ref{eqC}  express the same meaning. That is to say, in the steady state of static scattering, we can use the 3D reconstruction ability of the neural radiation field to fit a stable haze space where light and particles do not change.

Then the voxels with low transparency in the space are the regions of haze after stabilization in the real scattering model.We can remove volume fog by removing some voxels. Then the Eq.\ref{eqnerf} can be rewritten as Eq.\ref{eqnerf_dehaze}. We just need to remove the voxels in the space that are smaller than the alpha threshold to achieve accurate non-prior image dehazing.

\begin{equation}
F_{\text { dehaze }}:[(x, y, z),(\theta, \psi)\rightarrow[(r, g, b), (\alpha>\Delta t)]
\label{eqnerf_dehaze}
\end{equation}

\subsection{Determining the dehazing threshold (alpha)}

Not all voxels are objects or haze due to the accuracy of sampling,  some voxels are frequently sampled to the boundaries of haze and objects.  As seen in Fig ~\ref{fig_p1bestssim}, there are some voxels of intermediate grey voxels among the mostly light grey and black voxels. However, for the entire voxel space, this portion of the boundary is quite small.There will be three phases when the threshold starts to change: the removal of a large amount of haze, the removal of a small amount of haze and a small amount of object details, and the removal of a large amount of object details. So long as we capture the threshold in the middle phases, we can successfully remove the haze and leaves a significant amount of object details.

We need to find the threshold at which the rendered image results are relatively slow to change when the voxel changes. We will verify the feasibility of PSNR and SSIM to evaluate image changes in experiments.

\subsection{Global and Non-global Haze}

Global haze and non-global haze are two different types of haze. The fundamental criterion for evaluation is if there is haze in the image's main subject. In general, non-global haze is not uniform. For instance, photos of the nighttime fog frequently have significant haze in the area where the lights are visible. There are two types of global haze: uniform and non-uniform. globally consistent haze, such as in photos from distant sensing or regular ground fog.  

In the global haze scene, the light often passes through the haze before illuminating the object, and then passes through the haze again before entering the camera.
In the non-global haze scene, the light irradiates the object first, and then enters the camera after a haze.
Compared with the non-global haze scene, the global scene light passes through the haze scene more times, and this conclusion will affect the subsequent algorithm pipeline.

So we need to double the difference between the rendered dehaze image and the hazy image to get an image where the light hits the object and bounces into the camera without passing through the haze.

It should be noted that the incoming light passing through the haze is a macro concept, and the light entering and exiting between different voxels in the rendering and scattering equations above is a microscopic process.

\subsection{Systom overview and Algorithm structure}

We developed a hardware and software-integrated haze removal system. Haze Image Take, Rebuild and Remove Voxels, and Post Processing are the three modules that make up the system.The specific details are shown in the Fig..
Either our camera array or the movement of  a single camera to capture images of the haze. Then, using a number of images, the hazy 3D voxel space is recreated. We also choose an appropriate threshold to eliminate the voxels that are thought to be hazy in this area. By using the image's change information, we limit the threshold. To obtain the required haze-free image, we re-render the 3D space with the haze voxels removed. In order to enhance the quality of rendered haze-free images due to various haze causes and information loss during rendering, we select specific image post-processing techniques.
Image post-processing including image super-resolution and dehazing optimization for uniform and non-uniform haze scenes.
Because our image dahazing processing is more accurate and closely resembles the real haze scene, it produces processing results that are more precise, the algorithm is more reliable, and it does not need prior knowledge or data-driven training.

\section{DATA PREPARATION}

\begin{figure*}
	\centering
    \centerline{\includegraphics[width=1\textwidth]{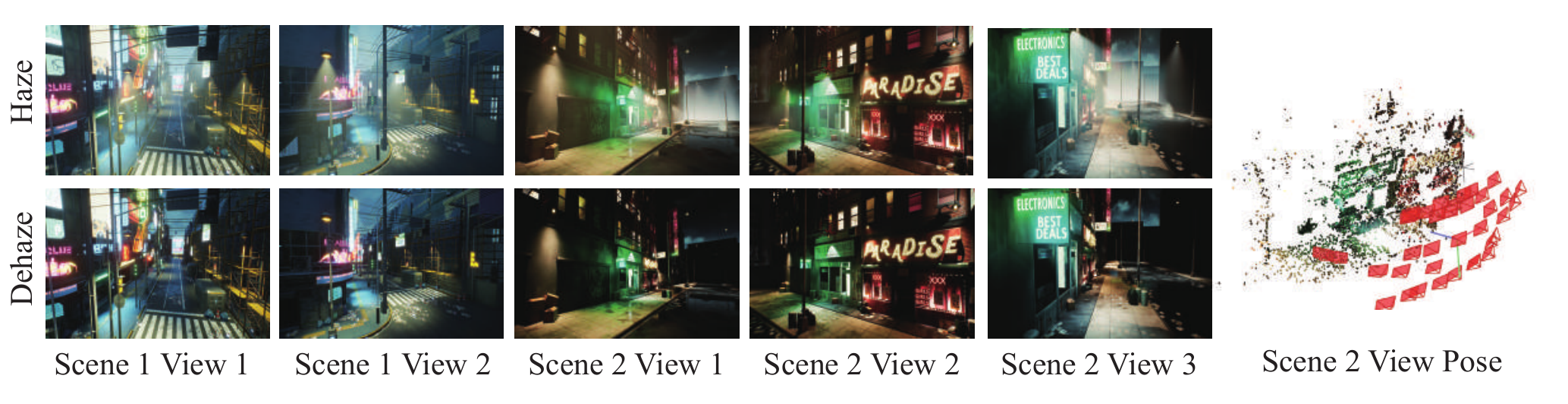}}
    \caption{\textbf{Simulate Data by Rendering Engine} We get the simulated array image through the rendering engine. Provide sufficient and reliable data for subsequent experiments. The rendering engine's volumetric fog algorithm is mature and conforms to the scattering formula. Moving objects and lights in the rendering engine can be turned off, and the haze color is determined by the light source. The haze can be adjusted in intensity or even turned off.}
    \label{fig_p4ue}
\end{figure*}
\subsection{Simulate Data by Rendering Engine} 

Existing image defogging data often only have two types: foggy and non-fog. However, we take mist, medium fog, thick fog images in read world, different images tend to have different fog densities. And it is difficulties to define the concentration of fog. So single-density haze data set has poor universality. However, our algorithm requires a haze simulate tool that can generate various concentrations of haze intensities and conform to the scattering formula like VRE or RTE.
 
Outdoor real-shot data set needs to consider issues such as object motion, light changes, haze floating and fog machine power. So it is very difficult to obtain real haze and ground truth image pairs that keep all other outdoor conditions. We proposed a data construction method based on Unreal Engine 5. The rendered image is much higher than the sub-pixel precision used in image processing. It is also much higher than the precision needed to reconstruct dehazing. 

We were able to gather data pairs for two different sorts of scenes uniform and non-uniform haze through the unreal engine 5  \cite{UE5}.

Many indicators in computer vision correspond to the field of image defogging. For example, Atmospheric Fog and Volumetric Fog correspond to Atmospheric haze and Glow Haze. Fog Density, Fog Falloff, Fog Scattering Color, Scattering Distribution, Albedo correspond to Optical Thickness, Attenuation Factor, Light source color, Atmospheric PSF, Glow haze gradient. Compared with two-dimensional image defogging, the three-dimensional parameters of computer vision fog effect are more complicated, There are many parameters that are not corresponding, such as Cast Volumetric Shadow and Volumetric Scattering Intensity. The simple correspondence is shown in the lower left corner of Fig ~\ref{fig_p4ue}.
 There are two main types of fog in the engine: Exponential Height Fog and atmospheric fog  \cite{UE5fog}.
Respectively represent the Glow and Airlight term in the image dehaze model. The main difference between night defogging and ordinary defogging is the volumetric fog which in the exponential height fog.

We move the virtual camera in the rendering engine in three-dimensional space, and fix the camera at a certain position. We can shoot foggy and non-fog scenes separately to obtain paired images. The image will be saved directly in the project folder. And because it is a virtual camera, the obtained images are completely consistent position, so subsequent operations such as registration are not required. It should be noted that some scenes have moving parts and special effects when shooting images. To keep the scene consistent, the moving parts and special effects need to be deleted. We can select different scenes and different views, as shown ~\ref{fig_p4ue}. The view pose result in \ref{fig_p4ue} is generated by Colmap. We finally produced 180 pairs of 3000*1600 pixel night fog and non-fog image pairs in different scenes.

Enlarging the image, it can be found that the collected image has the characteristics of directional haze, atmospheric light  and original image signal attenuation. The specific details are shown in \ref{fig_p4ue}. Even we found the rendered scene also shows that haze reduced the contrast of shadows. These characteristic changes are often lack of consideration in ordinary image defogging.

\begin{figure*}
	\centering
    \centerline{\includegraphics[width=1\textwidth]{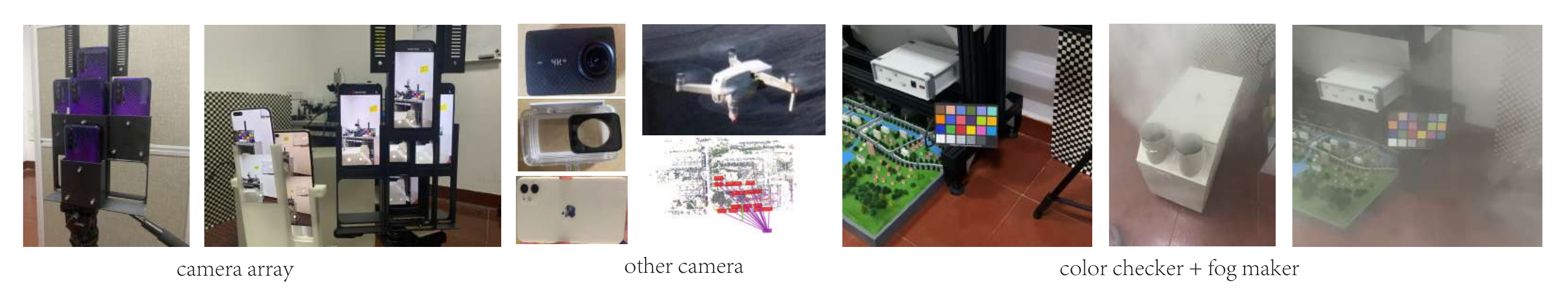}}
    \caption{\textbf{various systems for taking haze images} In order to collect the data for typical ground scenes, we use a camera array. The camera array mount is 3D printed in resin. When using camera arrays, issues like haze drifting are not necessary to take into account. We use drone cameras, action cameras with waterproof housings, and iPhones to shoot scenes with relatively fixed haze, such as remote sensing, underwater, at night, etc. We create haze for indoor scenes using a 600w ultrasonic fog machine. We also included a 24-color checker card in some data sets to check the camera's color accuracy and see how dehazing affected color recovery.}
    \label{fig_p6camera}
\end{figure*} 

\subsection{Array Camera and Fog Maker} 
For the actual data collection, we shot each scenario using a different way, as shown in Fig~\ref{fig_p6camera}.

We waited for the water and things to stay steady while rotating the waterproof camera back and forth to snap underwater pictures.

We use a moving drone camera that is pointed at the ground to capture images for remote sensing.

We use an iPhone to record video during foggy days to get non-uniform haze visuals at night.

We create haze for indoor photographs using a 600w ultrasonic fog machine. The floating haze will have an impact on the algorithm, but the device's video recording technology is unable to capture the information about the haze in real time. So, to obtain the photos, we employ a modest camera array. There are a minimum of five Android devices that are all the same and use fixed focus recording. The set up is as depicted in the \ref{fig_p6camera}. Video timestamps are used to synchronize the time.

\begin{table*}  
  \caption{ Quantitative Comparison for the impact of various image number on the results}
\centering	
\begin{tabular}{lccccccccc}  
\toprule   
 Number  &4 & 6 & 8 & 10 & 12 & 14 & 16 & 18 & 20  \\
\midrule   
 
PSNR-haze  &16.58  & 17.32 & 19.49 & 20.83 & 21.14 & 21.23 &21.10 & 21.06 & 21.32\\
SSIM-haze  &0.5388 & 0.5680 &0.7184 & 0.7370 & 0.7442 & 0.7535 & 0.7546 & 0.7542 & 0.7536\\
PSNR-dehaze  &15.10 & 15.59  & 17.91 & 19.43 & 19.46 & 20.26 & 20.36 &20.46& 20.52\\
SSIM-dehaze  &0.4360  & 0.4333 &  0.6122 & 0.6223 & 0.6361 & 0.6457 & 0.6517 & 0.6528 & 0.6621\\
 
  \bottomrule  
\end{tabular}

  \label{tab_number}
\end{table*}

\subsection{foggy image registration}
Reconstruction of haze space requires registration of hazy images first.

For new data, the method in this paper should first use the colmap algorithm to calculate. The aforementioned camera array method can also use the previously calibrated camera parameters.

As the haze density increases, the features in the image become more and more blurred, and the image registration becomes more and more difficult. In our tests on simulated data, when the foggy image density (voxel) a is greater than 0.2, the success rate of image matching drops greatly and the dataset is almost unusable.

Fortunately, in the long-term data collection process of haze days, it is rare to encounter global haze and the calculated voxel a is greater than 0.2. Most of the data concentrations are distributed between 0.01-0.1.There are almost no real images where the haze concentration is too high to cause the images to fail to match.

\section{Experimental Assessment}
To evaluate our technique, we carry out a comprehensive set of experiments aiming to answer the following three questions:

\begin{itemize}
\item[$\bullet$] 
What are the advantages of the proposed algorithm compared to existing dehazing algorithms.
\end{itemize} 
\begin{itemize}
\item[$\bullet$] 
Robustness of the algorithm under different haze concentrations, and the effect of the number of angles on haze removal.
\end{itemize} 
\begin{itemize}
\item[$\bullet$] 
Whether the algorithm is always applicable in each dehazing scene. How does it affect the algorithm in different real shooting data scenarios.

\end{itemize} 
\subsection{The Appropriate Dehazing Threshold}

\begin{figure}[h]
  \centering
  \includegraphics[width=\linewidth]{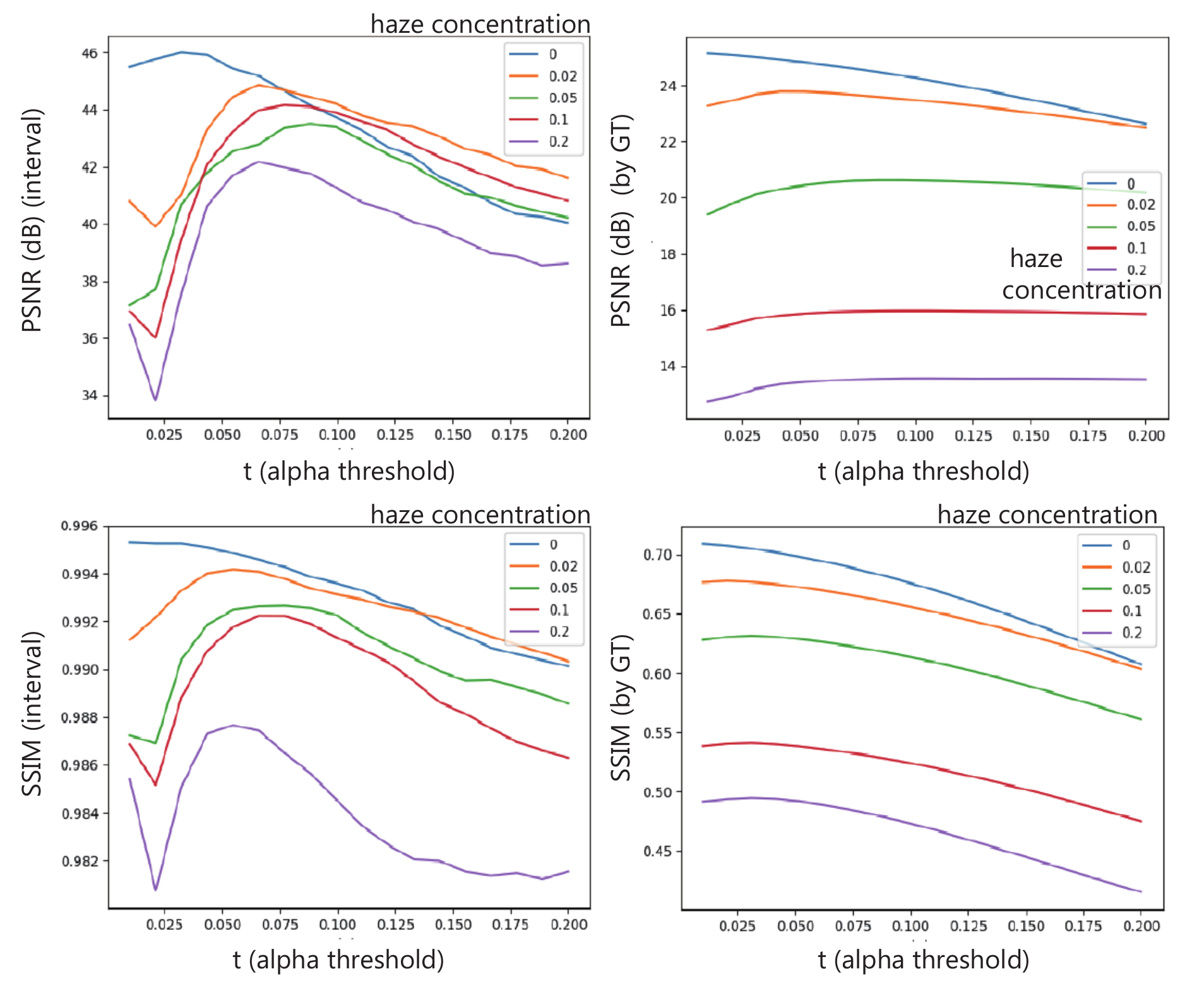}
  \caption{\textbf{Threshold voxel deletion experiment} The threshold method's viability is shown by the experimental results. Different haze concentrations are represented by lines of varying colors from 0 to 0.2. Images from thresholds t and t+$\Delta$t are compared in intervals psnr and ssim. Images and haze-free images are compared in psnr and ssim by GT at various t thresholds. The best image dehazing effect that is the highest psnr and ssim in comparison to the haze-free reference image occurs where the threshold change has the least impact on the image. }
  \label{fig_p1ssim}
\end{figure}

The threshold method's viability is shown by the experimental results in Fig ~\ref{fig_p1ssim}. Different haze concentrations are represented by lines of varying colors from 0 to 0.2. Images from thresholds $t$ and $t+\Delta t$ are compared in intervals psnr and ssim. Images and haze-free images are compared in psnr and ssim by GT at various t thresholds. The best image dehazing effect that is the highest psnr and ssim in comparison to the haze-free reference image occurs where the threshold change has the least impact on the image.

\subsection{Number of Reconstructed Images}
With or without haze, the number of reconstructed images often affects the quality of 3D neural radiation field reconstructions. 

The computational accuracy of the entire voxel (haze particles) is limited by the number of sampled images and the discrete sampling accuracy of the network. The higher the network accuracy, the higher the computational cost.

Many algorithms are being improved for small numbers of images, but these algorithms often require encoding. The premise of spatial coding is that there is no haze in the space, so we do not use spatial coding in our algorithm.

As shown in the Table\ref{tab_number}, the reconstruction effect of 4 to 10 haze increases rapidly. From 10 to 20 shots, the effect is slightly elevated, and sometimes noise is present. However, the haze removal effect of 4 to 10 pictures also increased rapidly, and the growth rate was more obvious than that of haze reconstruction. The growth of 10 to 20 pictures slowed down, but it was also more obvious than that of haze reconstruction.
\begin{figure*}
	\centering
    \centerline{\includegraphics[width=1\textwidth]{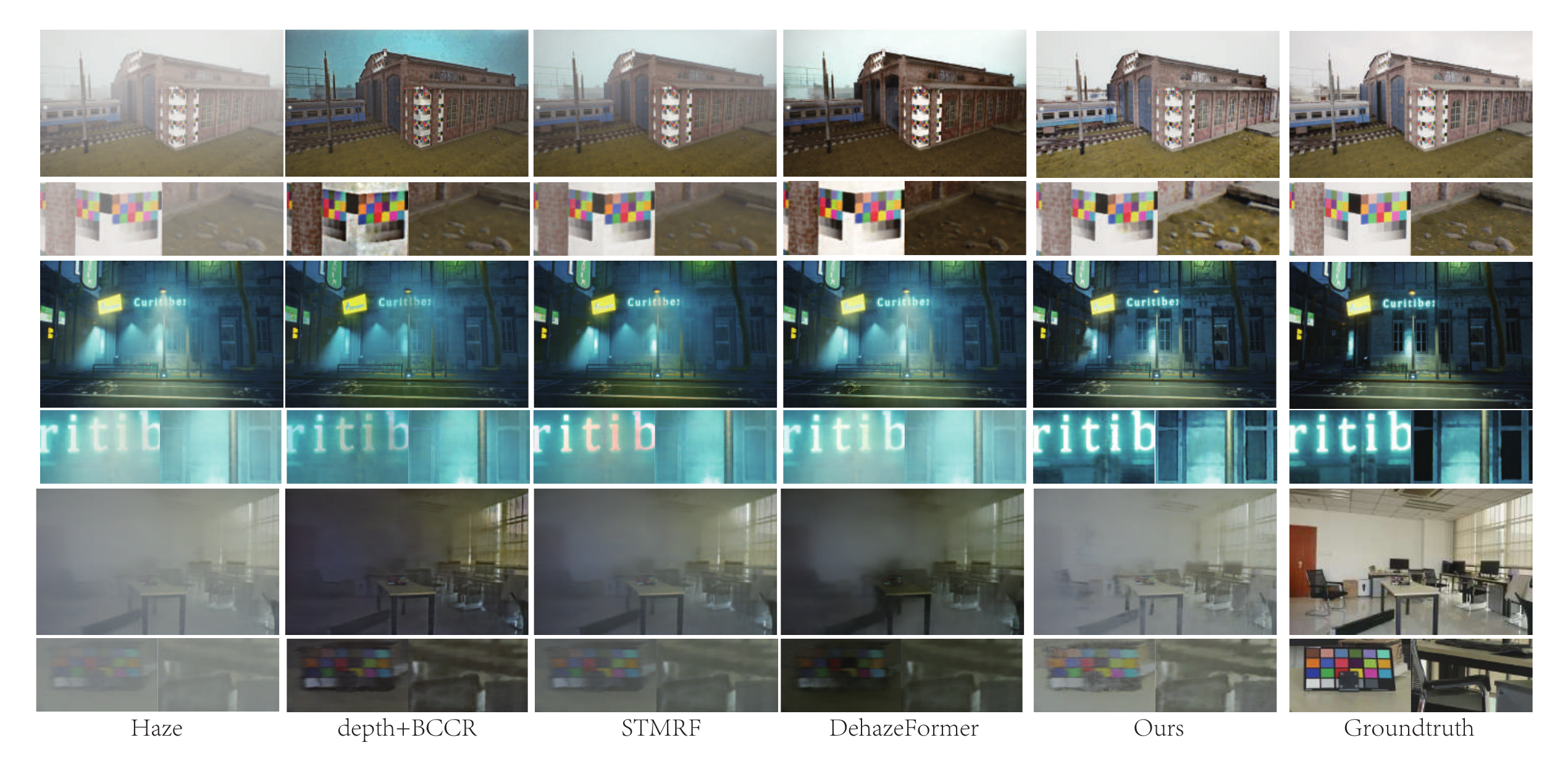}}
    \caption{\textbf{Hazy image dehazing result with ground truth.} The comparison of dehazing outcomes between uniform and non-uniform simulation datasets is shown in the first and second rows. It is evident that our method produces results that are more similar to actual haze-free scenes, richer picture details, all of the haze removed from them, and have more consistent color. The third layer results are those from the publicly available dataset; the haze there has moved, so our method is unable to fully remove it. However, it still outperforms other approaches and exhibits better color reproduction. A zoom-in of some details can be seen in the small image below.}
    \label{fig_p2result_gt}
\end{figure*} 

\begin{table*} 
 \caption{Quantitative Comparison for the differernt NeRF methods}
 \label{tab_nerfadd}
\centering	
\begin{tabular}{lccccc}  
\toprule   
 Method  & NeRF & NeRF+NGP16 & NeRF+NGP32  &  NeRF+NGP+Pre & NeRF+NGP+CUDA  \\  
\midrule   
non-uniform  &20.26 / 0.7862   &20.10 / 0.7438 & 20.11 / 0.7452 & 20.10 / 0.7445 & 20.06 / 0.7392\\
non-uniform(dehaze) &20.98 / 0.7646  & 17.31 / 0.6634  & 17.35 / 0.6652 & 17.56 / 0.6159 & 17.16 / 0.6718 \\
 
uniform       & 37.23 / 0.9220 & 36.18 / 0.9118 &36.19 / 0.9121 & 36.20 / 0.9119 &35.87 / 0.9007   \\
uniform(dehaze) &20.74 / 0.7918 &19.32 / 0.7679 &19.87 / 0.7711 &19.18 / 0.7673 &14.90 / 0.6643 \\

nohaze     & 39.89 / 0.8771   & 39.59 / 0.9382 & 39.58 / 0.9375 & 39.60 / 0.9388 & 39.49 / 0.9387 \\

nohaze(dehaze)   & 38.17 / 0.9886 & 38.94 / 0.9492 & 38.98 / 0.9537 & 38.93 / 0.9507 &39.54 / 0.9892     \\

  \bottomrule  
\end{tabular}
\end{table*}

\subsection{Comparison of different reconstruction algorithms}

Nerf-style papers are developing quickly, and focus on greater texture details and faster calculation. Pay attention to high-alpha regions of space that include opaque objects. In the reconstruction space, low-weight foggy regions are frequently disregarded. We tested various class nerfs and contrasted them.
 
When there is no haze in the image, there are a large number of transparent and unimportant voxels in the three-dimensional space.
When there is haze in the image, the voxels in the 3D space can no longer be simply encoded and omitted, and each voxel needs to be accurately recorded for correct dehazing. Our experimental results also show similar results in Table \ref{tab_nerfadd}.

Compared to other methods, the basic nerf \cite{NeRF}is the slowest, the most accurate and the most stable.
Other methods use different precision encoding, data preloading, and cuda acceleration methods.

\begin{table*} 
 \caption{Quantitative Comparison for the differernt methods on real images}
 \label{tab_result}
\centering	
\begin{tabular}{lccccc}  
\toprule   
 Method  & Haze & depth+BCCR & STMRF  & DehazeFormer  & Ours  \\  
\midrule   
non-uniform real     & 19.18 / 0.4348 &18.85 / 0.4412 & 15.74 / 0.4438 & 18.93 / 0.4509 &   \textbf{26.02} /   \textbf{0.7152}  \\
non-uniform unreal   & 13.19 / 0.6229 & 15.54 / 0.6326 & 16.10 / 0.6491 & 14.82 / 0.6482 &  \textbf{20.98} /  \textbf{0.7646}  \\
uniform real      & 15.28 / 0.5668 & 10.09 / 0.4653 & 13.79 / 0.5560 & 10.90 / 0.4966 &  \textbf{15.83} /  \textbf{0.5745}   \\
uniform unreal    & 14.36 / 0.6387 & 12.79 / 0.6500 &   \textbf{20.84} / 0.7227 & 14.56 / 0.6564 & 20.74 /  \textbf{0.7918}  \\
nohaze real    & $ \infty $  /  1 & 12.28 / 0.8415 & 14.64 / 0.8095 & 16.56 / 0.8521 &  \textbf{25.43} /  \textbf{0.8578} \\
nohaze unreal  & $ \infty $  /  1 & 22.62 / 0.9625 & 21.77 / 0.8372 & 25.87 / 0.8660 &  \textbf{40.17}  /   \textbf{0.9886}  \\
  \bottomrule  
\end{tabular}
\end{table*}

These experiments test the haze reconstruction ability and haze removal ability (dehaze) of different algorithms under uniform haze, non-uniform haze and no haze scenarios.Compare and evaluate the reconstructed and rendered image of the scene with the ground truth. 

The ability to reconstruct haze scenes, the gap between various algorithms is not large, the native NeRF effect is slightly better. NGP+CUDA  \cite{CUDA,NGP} has a huge advantage in speed, so result is a little poor. There will also be a little improvement in using a higher precision encoding,such NGP32 always a little better than NGP16.

There is a big gap between different algorithms in terms of haze removal ability, and the size of the gap is also related to the scene.

\begin{figure*}
	\centering
    \centerline{\includegraphics[width=1\textwidth]{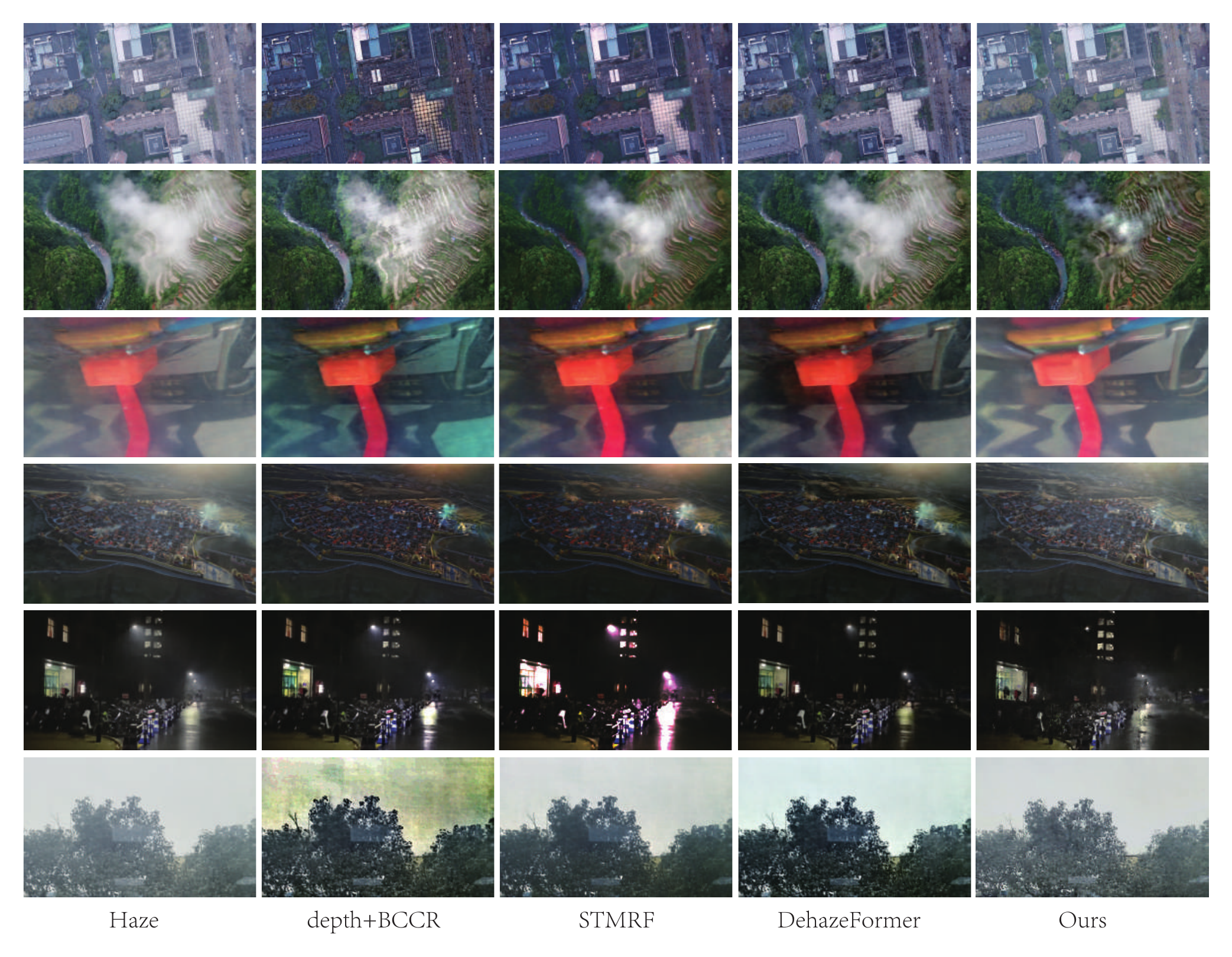}}
    \caption{\textbf{Real Hazy image dehazing result without ground truth.}There are usually no fog-free images to compare to foggy array images. Our algorithm can effectively maintain the true color of the image even in the presence of remote sensing, underwater, glass reflections, drones, nighttime and non-uniform ground haze. Other algorithms always distort colors and remove too much haze. Compared with the simulated images, our method also has a certain degenerate in the real-shot images, mainly due to the floating problem of haze and the scarcity of the number of reconstructed images.}
    \label{fig_p2result}
\end{figure*} 
For haze-free scenes dehazing, the image needs to remains. the NeRF use of encoding and accelerated algorithms is often better,  because the voxel calculation is often more compressed, and the image degradation in the voxel removal is smaller.  Simple NeRF will have more wrong voxel removal part.

For non-uniform haze scenes, encoding is the decisive factor for the ability to dehaze. Once encoding is started, the non-uniform haze will lose its transparency and become a part of non-transparent voxels, which cannot be distinguished in the process of dehazing. Only traditional NeRF can handle non-uniform haze robustly. During the experiment, we found that the NGP type method can also deal with some non-uniform haze, and the results can also exceed the traditional NeRF, but the complete failure in some data reduces the results of the whole dataset.

For uniform haze scenes, both NGP and traditional nerf can perform haze removal well, and the performance gap is not large, but accelerated CUDA will encode haze voxels incorrectly.

If don’t care about memory and time overhead, traditional NeRF is the best. If you can know the type of scene in advance, using NGP and other encoding and CUDA acceleration can also ensure the quality of dehazing.

\subsection{Synthetic Image Dehazing}
Qualitative Comparison: The simulation data does not take into account the movement of real-world haze and the error of camera imaging. Therefore, our algorithm achieves very good results in Fig. ~\ref{fig_p2result_gt}.
For other algorithms, even if the scene depth information is collected or based on the most advanced network structure, there is no way to ensure the accuracy of the color, and there is often a serious color cast.Our algorithm can almost match the ground truth visually. The contrast of color checker and texture details is far superior to other algorithms.
For other non-uniform fog removal scenes such as nighttime fog removal, other methods have almost no ability to remove fog. Other algorithms can fit part of the haze through depth information and the network, but cannot know the specific intensity of the haze.On the contrary, our algorithm can excellently suppress the haze and scattered glow caused by the light source, and restore the details under the haze.

Quantitative Comparison:For almost all simulation scenarios, whether it is homogeneous haze or non-uniform haze, or a haze-free scenario that reflects the robustness of the algorithm, our algorithm shows absolute leadership in quantitative analysis.The specific data is shown in Table ~\ref{tab_result}. When the depth is known, STMRF can slightly lead our method based on the depth information, but the image dehazing result is obviously color cast and the dehazing is insufficient.

\subsection{Real Image Dehazing}
The biggest problem with real images compared to simulated images is that outdoor scenes are difficult to achieve haze and stillness of the scenery.
In Fig.~\ref{fig_p2result_gt}, we used foggy images taken by others using a robotic arm \cite{hazedata}, and we found that our algorithm could not take absolute advantage due to the haze movement during the robotic arm movement. However, it is still ahead of other algorithms in terms of color retention and image texture.

We compare the results of different algorithms on our real-world data, as the real-world images are not available for reference as the ground truth images.
The remote sensing image, the non-uniform haze drone image, the underwater image, the local non-uniform haze image with insufficient light, the night haze image and the window reflection image are shown in Fig. \ref{fig_p2result}.

Compared with the simulated images, our method also has a certain degenerate in the real-shot images, mainly due to the floating problem of haze and the scarcity of the number of reconstructed images.It's important to note that all of the result data here require less than 10 images to reconstruction.

Our results work well in every scene. The remote sensing images are more able to recover the real colors and features, and the road and vegetation colors are more accurate and not oversaturated like other algorithms. The drone images can evenly distinguish haze and remove it. And it can maintain the correct white balance. Underwater images remove turbidity from water and maintain correct color. Nighttime images suppress non-uniform directional haze and maintain slight color correction of lights. Reflection ghosting is also removed and color and texture details are maintained in image de-reflection.

Some algorithms can sense where there is haze based on the depth information of the image dataset, but it is not clear how much haze has to be removed and what the true color of the haze is after removal. Deep learning-based algorithms, even if they are as strong as the latest transformer-based networks and use the corresponding image dataset, can obtain better results than algorithms that simply know the depth location information, but they are still limited by the a priori knowledge of the dataset \cite{dehazeformer}. For example, the specially optimized nighttime defogging achieves good results, but is still inferior to our algorithm. Other domains have problems such as color bias and transition defogging once images unfamiliar to the network appear.

\begin{figure*}
	\centering
    \centerline{\includegraphics[width=1\textwidth]{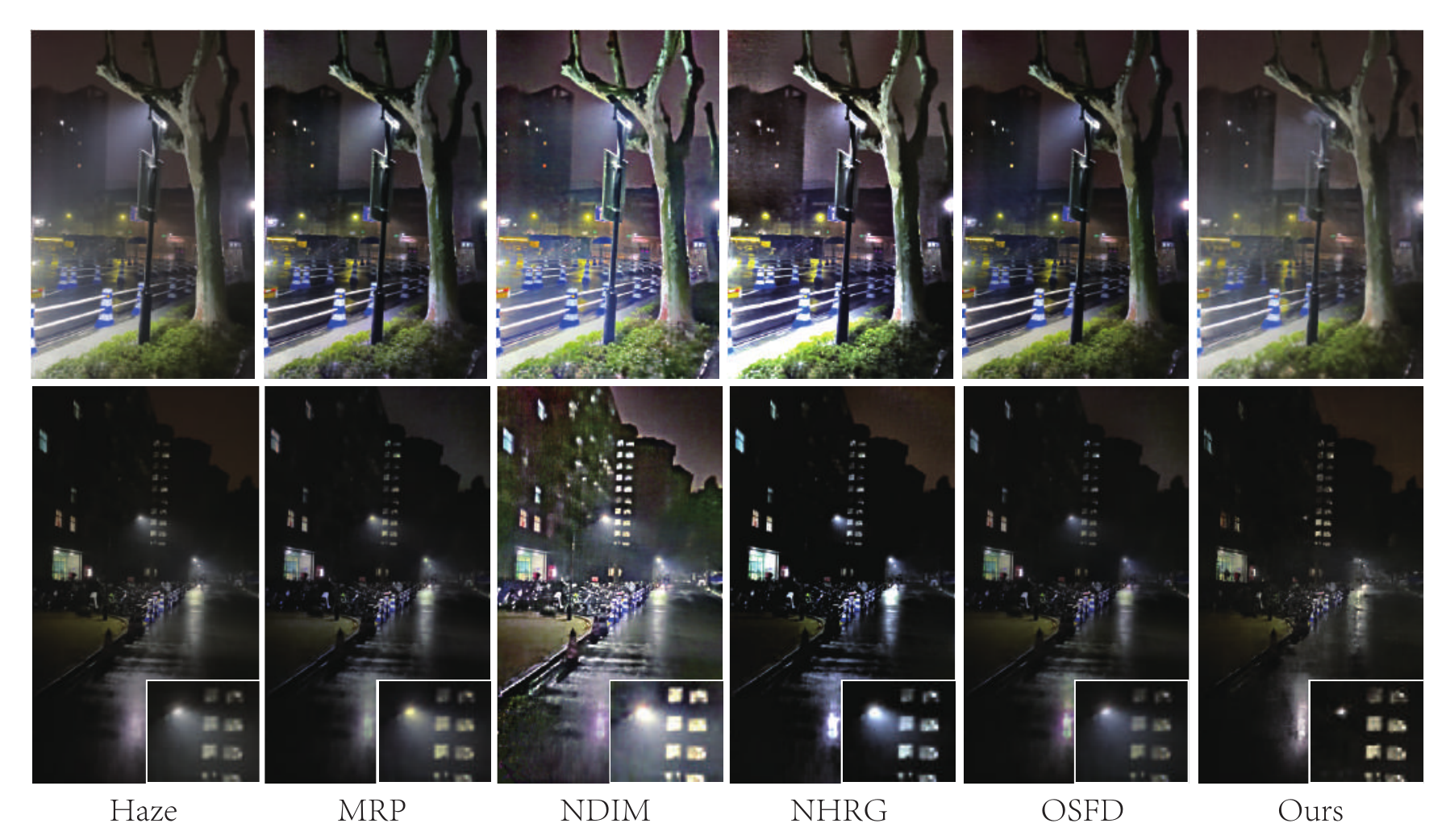}}
    \caption{\textbf{Real Nighttime Hazy image dehazing result.} Our method done well of removing haze caused by directional light sources and glare near lights. And still maintain the normal color of the scene and the light in the ground specular reflection and diffuse reflection. }
    \label{fig_p2result_nighttime}
\end{figure*}

To show our strengths in every single area, we chose the most difficult of nighttime dehazing for comparison. The color and brightness
 of the nighttime defogging light sources are variable and the haze is non-uniform, and there is a lack of publicly available and reliable data sets. A comparison of our method with several recent nighttime dehazing method is shown in Tab.~\ref{tab_nighttime_result} and Fig.~\ref{fig_p2result_nighttime}, and it is clear that our method perceives the location and intensity of the haze accurately and has good dehazing effect. Other algorithms do not work well even if they use special dark nighttime dehazing priors and specially designed datasets.
 
\begin{table*} 
 \caption{Quantitative Comparison for the differernt methods on nighttime images}
 \label{tab_nighttime_result}
\centering	
\begin{tabular}{lcccccc}  
\toprule   
 Method  &Haze &NDIM&NHRG & MRP & OSFD & Ours   \\  
\midrule   
PSNR SSIM$\uparrow$  &16.17 / 0.5677 & 10.56 / 0.3102 &13.95 / 0.4842  &16.43 / 0.5220  &17.72 / 0.5674 & \textbf{ 20.44 / 0.7644}  \\  
CIE2000$\downarrow$ &146.28   &130.08 &118.69  &132.59  &128.87  & \textbf{103.32 }  \\
  \bottomrule  
\end{tabular}
\end{table*}

\section{CONCLUSION, LIMITATIONS, AND FUTURE WORK}

Our work proposes new ideas for future high-performance, high-confidence, and robust image dehazing, combining computer rendering, image dehazing, and 3D reconstruction.
Exciting results have been achieved on both simulated and real-shot images. OUr algorithm is very accurate and robust, it can be used in various fields, and does not require any data and training process

Many people will think that the baseline between different cameras and the resolution of the camera for the dehazing method in this paper are very important.
In our tests, this was not the deciding element. Theoretically, the camera parallax should have less of an impact the further it is away, but the reconstructed camera rays will also be illuminated more.

In actuality, the interaction between the haze and other scenes has a significant impact on the algorithm's effectiveness.
Our methods use the variations brought on by haze on the same object when captured by various cameras. The result is better if the background has a complex texture.

We have demonstrated the benefits of dehazing by NeRF directly. However, this modification is not without tradeoffs. It requires a camera array, which is easy to obtain in remote sensing satellites, autonomous driving, and scientific exploration. , but in personal consumer goods such as SLR collection, it is often necessary to shoot video, and shooting video often encounters objects and haze movement, etc. Like  \cite{2021hNeRF}, our method is also dependent on COLMAP’s robustness for computing camera poses, preventing us from capturing scenes above a certain haze level. This could potentially be addressed by dehazing method and the input camera poses. 

Finally, despite its robustness to haze, our method cannot be considered a general purpose dehazing. As it cannot handle fog motion and requires orders of magnitude more computation than a feed-forward network.

Despite these shortcomings, we believe that ourmethod 
represents a step toward robust, high quality capture of real world environments haze.  Lifting these constraints
greatly increases the fraction of the haze that can be reconstructed and explored with photorealistic view synthesis.

Our approach may be extended in several areas:
The first is exact encoding for non-transparent voxels. Although the threshold discrimination method used in this paper is simple to implement and easy to understand, the effect still has room for improvement. The existing 3D reconstruction uses coding methods to enhance the network's perception of spatial voxel changes. We believe that semi-transparent voxels can also be used. A similar encoding method is proposed to enhance the network's understanding of the haze space, thereby improving the accuracy of haze removal.

The second large number of excellent image dehazing datasets are on the horizon. The existing outdoor large scene datasets are difficult to obtain. After applying the method in this paper, as long as there is an array camera, the outdoor dehazing image pairs can be accurately obtained.

In addition to the real shot data set, haze simulation has also entered a new height. With the rapid development of 3D reconstruction, a large amount of trained or untrained data can be turned into a foggy 3D space by increasing the weight alpha , so as to output a large number of fog-free, uniform and non-uniform haze datasets in the same scene.

Third, the expansion of the field of image dehazing. Although this article has tried some remote sensing, the method of underwater night dehazing has also achieved certain results, but there are still some differences between different dehazing methods. There are also many general methods in this article. Optimize space. For example, there are also directions such as multispectral dehazing, etc., which can also be applied to the method in this paper.

\begin{acks}

\end{acks}


\bibliographystyle{ACM-Reference-Format}

\bibliography{sample-base}

\appendix

\end{document}